\documentclass[10pt,twocolumn,letterpaper]{article}

\usepackage{cvpr}              

\usepackage[dvipsnames]{xcolor}

\usepackage{multirow}
\usepackage{mathtools}
\usepackage{breqn}
\usepackage[accsupp]{axessibility}

\newcommand{\Lplau}{{\mathcal{L}_{\text{\small \tt plau}}}}
\newcommand{\Lrep}{{\mathcal{L}_{\text{\small \tt rep}}}}

\newcommand{\visllm}{large video-language model }
\newcommand{\visllms}{large video-language models }
\newcommand{\Visllm}{Large Video-language model }

\newcommand{\epickitchens}{EPIC-Kitchens-100\xspace}
\newcommand{\method}{PlausiVL\xspace}

\definecolor{cvprblue}{rgb}{0.21,0.49,0.74}
\usepackage[pagebackref,breaklinks,colorlinks,citecolor=cvprblue]{hyperref}

\def\paperID{6111} 
\def\confName{CVPR}
\def\confYear{2024}

\title{Can’t make an Omelette without Breaking some Eggs: Plausible Action Anticipation using Large Video-Language Models}

\author{Himangi Mittal$^{1,2}$\thanks{\:This work was done as Himangi Mittal’s internship project at Honda Research Institute USA.} ~~~~~~~~ Nakul Agarwal$^1$ ~~~~~~~~ Shao-Yuan Lo$^1$ ~~~~~~~~ Kwonjoon Lee$^1$ \\
$^1$Honda Research Institute USA ~~~~ $^2$Carnegie Mellon University\\
{\tt\small hmittal@andrew.cmu.edu} ~~~
{\tt\small \{nakul\_agarwal, shao-yuan\_lo, kwonjoon\_lee\}@honda-ri.com}
}

\begin{document}
\maketitle
\begin{abstract}

We introduce PlausiVL, a \visllm for anticipating action sequences that are plausible in the real-world. While significant efforts have been made towards anticipating future actions, prior approaches do not take into account the aspect of plausibility in an action sequence. To address this limitation, we explore the generative capability of a \visllm in our work and further, develop the understanding of plausibility in an action sequence by introducing two objective functions, a counterfactual-based plausible action sequence learning loss and a long-horizon action repetition loss. We utilize temporal logical constraints as well as verb-noun action pair logical constraints to create implausible/counterfactual action sequences and use them to train the model with plausible action sequence learning loss. This loss helps the model to differentiate between plausible and not plausible action sequences and also helps the model to learn implicit temporal cues crucial for the task of action anticipation. The long-horizon action repetition loss puts a higher penalty on the actions that are more prone to repetition over a longer temporal window. With this penalization, the model is able to generate diverse, plausible action sequences. We evaluate our approach on two large-scale datasets, Ego4D and \epickitchens, and show improvements on the task of action anticipation.




\end{abstract}    
\section{Introduction}
\label{sec:intro}
\begin{figure}[t]
    \centering
    \includegraphics[trim=1.5cm 1cm 9.2cm 0cm, clip, width=\linewidth]{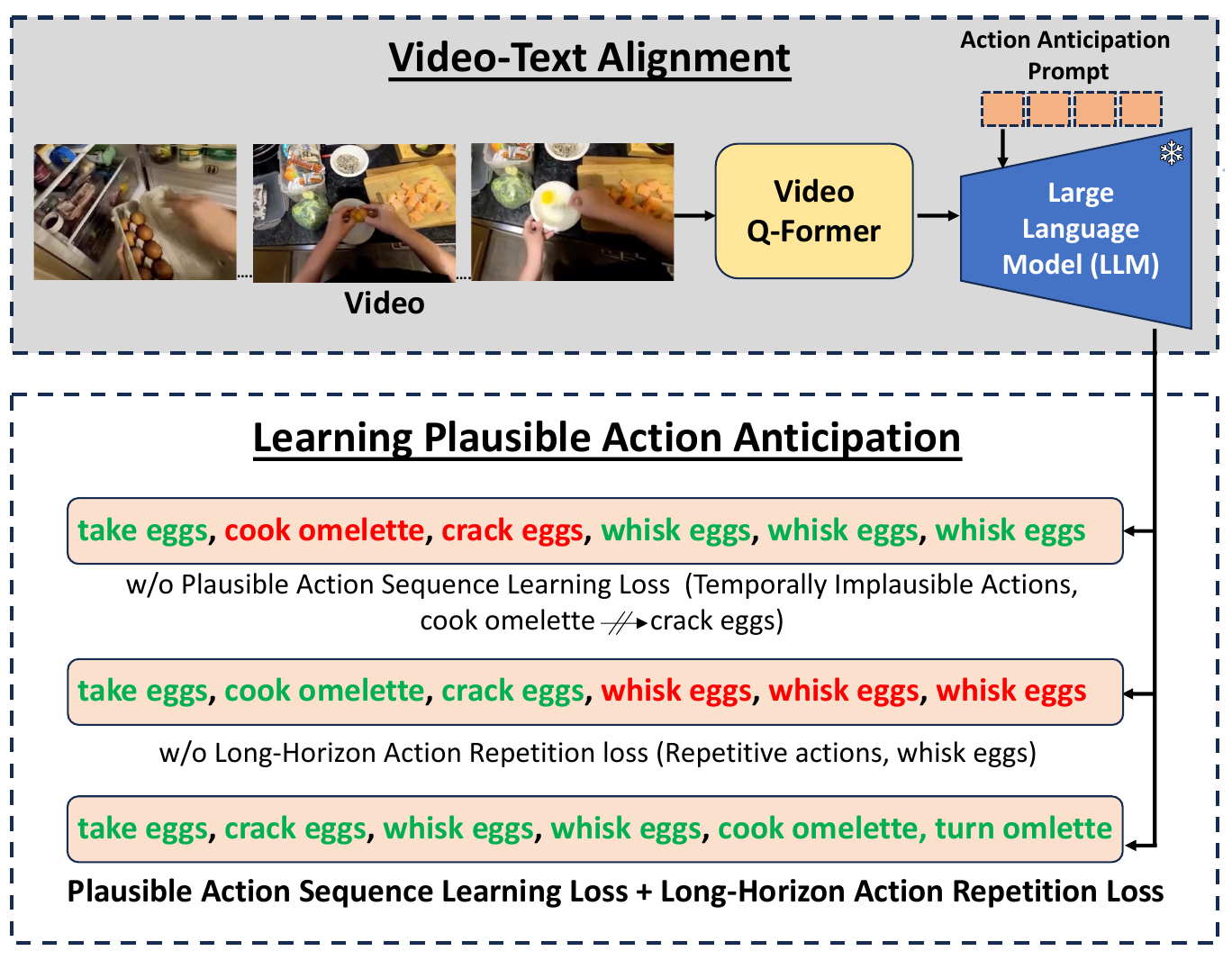}
    \caption{We present a \visllm for learning to anticipate action sequences that are \textit{plausible} in the real-world. We show an example of a kitchen-based environment. By using a \visllm, we leverage their generative capabilities to anticipate future actions and further train the model with two devised objective functions: plausible action sequence learning loss and long-horizon action repetition loss. Without the plausible action sequence learning loss, the model has less temporal understanding and generates a temporally implausible action sequence of \textcolor{red}{\textit{cook omlette} $\not \rightarrow$ \textit{crack eggs}}. Similarly, without the long-horizon action repetition loss, the model generates less diverse actions and repeats the same action, \textcolor{red}{\textit{whisk eggs} $\rightarrow$ \textit{whisk eggs} $\rightarrow$ \textit{whisk eggs}}. When training the model with the two objective functions combined, our method is able to generate plausible action sequences which are temporally accurate, \textcolor{Green}{\textit{crack eggs} $\rightarrow$ \textit{cook omlette}} and more diverse with less repetition, \textcolor{Green}{\textit{whisk eggs} $\rightarrow$ \textit{whisk eggs} $\rightarrow$ \textit{cook omlette}}.}
    \vspace{-4mm}
    \label{fig:intro_diagram}
\end{figure}

Having the ability to predict future events is a critical component in the decision-making process of an AI agent. For example, for an autonomous driving car, being able to anticipate the next sequence of actions for cars, pedestrians, and other agents in the scene can ensure safety of pedestrians as well as vehicles. To enable this, the model should be able to reason effectively from the spatial as well as temporal information of the visual scene. This has led to a growing interest in the task of \textit{Action Anticipation}. Action anticipation refers to the predictive task of forecasting future actions or activities given a sequence of visual data, typically videos. 
For example, in a kitchen-based environment, if a human has performed the following series of actions, \textit{open fridge} $\rightarrow$ \textit{take eggs} $\rightarrow$ \textit{close fridge}, the model should be able to reason that \textit{crack eggs} could be one of the plausible future actions.

However, action anticipation is challenging because the uncertainty in precisely predicting the future makes the task non-deterministic in nature. In other words, given what has happened so far, there are infinitely many possibilities for what future actions might happen. Moreover, action anticipation is accompanied by an additional challenge of understanding the implicit temporal information present in an action sequence, which makes the sequence \textit{plausible} in the real-world. For example, the model should be able to understand that an action like \textit{crack eggs} will always happen \textit{\textbf{before}} \textit{cook omelette} as shown in Figure~\ref{fig:intro_diagram}. 



To this end, there has been some progress for the action anticipation task. Earlier works have explored an LSTM based approach by summarizing the past and inferring the future~\cite{furnari2020rolling, osman2021slowfast}, by logging the past history actions in text~\cite{manousaki2023vlmah}, or using RNN-based approaches~\cite{roy2022predicting, shi2018action} by learning goals. However, such LSTM/RNN-based approaches are unable to effectively capture the temporal relations among the actions over a long horizon due to their sequential nature. Recent works have also explored transformer-based approaches~\cite{girdhar2021anticipative, gong2022future, roy2022interaction}, with a memory-based system~\cite{wu2022memvit} or leveraging multiple-modalities~\cite{zhong2023anticipative,zhang2023object}. While transformer-based approaches are able to model longer temporal understanding, they can still become confined to the information present in the training data and cannot model the diverse nature of the future actions. They rely on the ability of the transformer encoder to learn from the given training data which limits their generalization and scaling capability.



To overcome the above challenges, recent methods~\cite{li2023blip,alayrac2022flamingo,wang2023all,li2023lavender} have attempted to leverage the autoregressive text generation capabilities of generative large-language models~(LLMs) to improve generalizability for various vision tasks. Taking inspiration from these works and to address the challenges present in anticipating plausible actions, we introduce \textbf{\method}, \textbf{Plausi}ble action anticipation through a large \textbf{V}ideo-\textbf{L}anguage model. 

Given the generative capabilities of large language models, in this work, we introduce a video-large-language model which can efficiently model and leverage the temporal cues present in a video to generate plausible action sequences for the task of action anticipation. We use a Q-former~\cite{li2023blip} based transformer architecture to embed videos into spatio-temporal visual representations. This architecture ensures an effective alignment between the visual features and the desired text in the LLM embedding space. In addition to the alignment, we try to address the challenges that are specifically present in the task of action anticipation and thus, introduce a method with the following important characteristics: 1). The ability to understand the temporal correlations present among the actions in a sequence which in turn makes the action sequence temporally \textit{plausible} in the real-world, 2). Being able to model the diverse, possible actions that can happen in the future. For example, for the former characteristic, a model should follow a temporal constraint that \textit{an action X has to happen before for the action Y to happen} to make the sequence \textit{action X} $\rightarrow$ \textit{action Y} plausible in the real-world.

To build such temporal understanding required for generating plausible action sequences, we design a counterfactual-based plausible action sequence learning loss where we create temporal logic constraints and train the model to be able to differentiate between the plausible and not plausible action sequences. Additionally, we also use verb-noun action logical constraints to further improve the model's understanding about which verbs are possible with which nouns to create a plausible action in the real-world~(for example, cook spoon is not a plausible action). To our knowledge, the aspect of plausibility in generating an action sequence has not been explored for the task of action anticipation. While this loss is helpful for efficient temporal understanding, we also aim for the model to be able to understand the diverse nature of actions and generate plausible action sequences with less repeated actions as language models are prone to the issue of repetition. To resolve this, we devise a long-horizon action repetition loss where the later actions that are more prone to repetition have a higher penalty and the earlier, immediate actions have lower penalty. We summarize our contributions as follows:

\begin{enumerate}
    \item We present \method, a \visllm which leverages the spatial-temporal information present in videos for anticipating plausible future action sequences.
    \item To learn the temporal cues and understand the temporal dependencies among actions in a plausible sequence, we design a counterfactual-based plausible action sequence learning loss. We create temporal logic rules and verb-noun action pair logic constraints for the model to be able to understand plausibility in action sequences.
    \item To be able to generate less diverse future actions with less repetition, we devise a long-horizon action repetition loss by penalizing the longer-horizon actions more.  
\end{enumerate}

\section{Related Works}
\label{sec:related_works}

\textbf{Large Language Models.}
Language Modeling is a method to model the generative likelihood over the word token sequences and predict the probabilities of the next/future tokens. Large language models~(LLMs)~\cite{brown2020language,chowdhery2022palm,taylor2022galactica,touvron2023llama} are transformers with billions of parameters that have been trained on massive amounts of data and have shown impressive capabilities on the task of question-answering and chat-conversation with humans. Methods like in-context learning~\cite{brown2020language}, prompt tuning~\cite{white2023prompt}, chain-of-thought reasoning~\cite{wei2022chain}, and reinforcement learning with human feedback~\cite{ouyang2022training,christiano2017deep} have improved the language models to perform very well on few-shot tasks. While these models show great capabilities in understanding the input and solving complex tasks via text generation, these models can only understand the text modality and are at a loss of the rich information that is present in other modalities like video, audio. In our work, we utilize videos as input and learn from the visual and temporal information present in them.

\noindent \textbf{Large Vision-Language Models.} 
Recent strides in this domain have seen diverse pre-training methods leveraging extensive multimodal datasets driving the progress of large vision-language models. Some models~\cite{radford2021learning, jia2021scaling, furst2022cloob, li2021supervision, wang2023protege} merge visual and linguistic modalities by co-training text and image encoders using contrastive loss on large datasets containing image-caption pairs. Meanwhile, other approaches~\cite{alayrac2022flamingo, chen2022visualgpt} integrate visual input directly into language model decoders through a cross-attention mechanism, eschewing the use of images as additional prefixes. Another category of vision-language models~\cite{li2019visualbert, lu2019vilbert, chen2020uniter, singh2022flava, tan2019lxmert, li2020oscar} leverage Masked-Language Modeling (MLM) and Image-Text Matching (ITM) objectives to align image segments with text.
BLIP-2~\cite{li2023blip} was one of the works which proposed a Qformer-based method to ensure visual-text alignment. Since these works explore the image-text alignment, they are unable to model and understand the temporal information that is present in videos. There have been efforts towards video-text alignment by using a linear layer to project the video space to the LLMs textual space~\cite{chen2023videollm} in Video-LLM or by using a Q-former based module~\cite{zhang2023video} in Video-LLaMA. While these works explore video-text alignment, these models can be ineffective for the task of action anticipation as they do not understand the temporal correlations among the actions in a sequence.

\noindent \textbf{Temporal and symbolic logic reasoning.} Symbolic logic reasoning is a method to create a system of rules and symbols in the form of logical expressions. Temporal logic reasoning specifically designs logical expressions for representing and reasoning about time. Linear temporal logic~\cite{pnueli1977temporal}, metric temporal logic~\cite{montanari1996metric}, signal temporal logic~\cite{fainekos2009robustness}, and interval temporal logic~\cite{halpern1991propositional} are some methods for develop temporal logical rules. We take inspiration from the work DTL~\cite{xu2022don} to generate temporal logic rules and create counterfactual sequences of actions.

\noindent \textbf{Action Anticipation.}
This task has been explored for third-person videos~\cite{vondrick2016anticipating,abu2018will,gao2017red,chen2022gatehub,rizve2023pivotal} as well as egocentric videos~\cite{rodin2022untrimmed, qi2021self, girdhar2021anticipative, furnari2022towards, damen2018scaling, damen2020rescaling, grauman2022ego4d}. Standard approaches for this task can be divided into LSTM/RNN-based~\cite{shi2018action, damen2020rescaling} approaches and transformer-based approaches. LSTM-based approaches~\cite{furnari2020rolling, osman2021slowfast} mainly use a rolling LSTM to encode the observed video and store an updated summary. For inference, an unrolling LSTM is initialized with the hidden and cell state of the rolling LSTM to predict the next action. While LSTM/RNNs have shortcomings in modeling long-horizon temporal dependencies, some approaches mitigate this issue via goal-based learning~\cite{roy2022predicting}, diverse attention mechanism~\cite{gong2022future}, skip-connections~\cite{ke2019time}, message passing framework~\cite{tai2022unified}, memory-based modules~\cite{wu2022memvit, manousaki2023vlmah} or similarity metric~\cite{fernando2021anticipating}. Recent works have explored transformer-based~\cite{girdhar2021anticipative, grauman2022ego4d} approaches with global attention~\cite{gong2022future}, modelling apperance change in human-object interactions~\cite{roy2022interaction}, conditioning on intention~\cite{mascaro2023intention}, hierarchical feature aggregation~\cite{mascaro2023intention}. While most of the works explore it in a unimodal setting by using the visual modality, other works also present a multi-modal approach for this task by using optical flow~\cite{furnari2020rolling,osman2021slowfast}, object-based features~\cite{furnari2020rolling,osman2021slowfast,zhang2023object} or audio~\cite{mittal2022learning,zhong2023anticipative}. Other works explore uncertainty-based methods~\cite{furnari2018leveraging, abu2019uncertainty,guo2024uadt} and GAN-based approach~\cite{gammulle2019predicting}. We take inspiration from the object detection~\cite{lin2017focal} literature for the repetition loss. Concurrent to our work, there have been text-based LLM approaches~\cite{zhao2023antgpt,huang2023palm} which explore the task of action anticipation, however, they only operate in the textual space and lose the visual-temporal information present in video.

\section{Method}
\label{sec:method}
\begin{figure*}[t]
    \centering
    \includegraphics[trim=2.5cm 8cm 2.5cm 3cm, clip, width=0.9\linewidth]{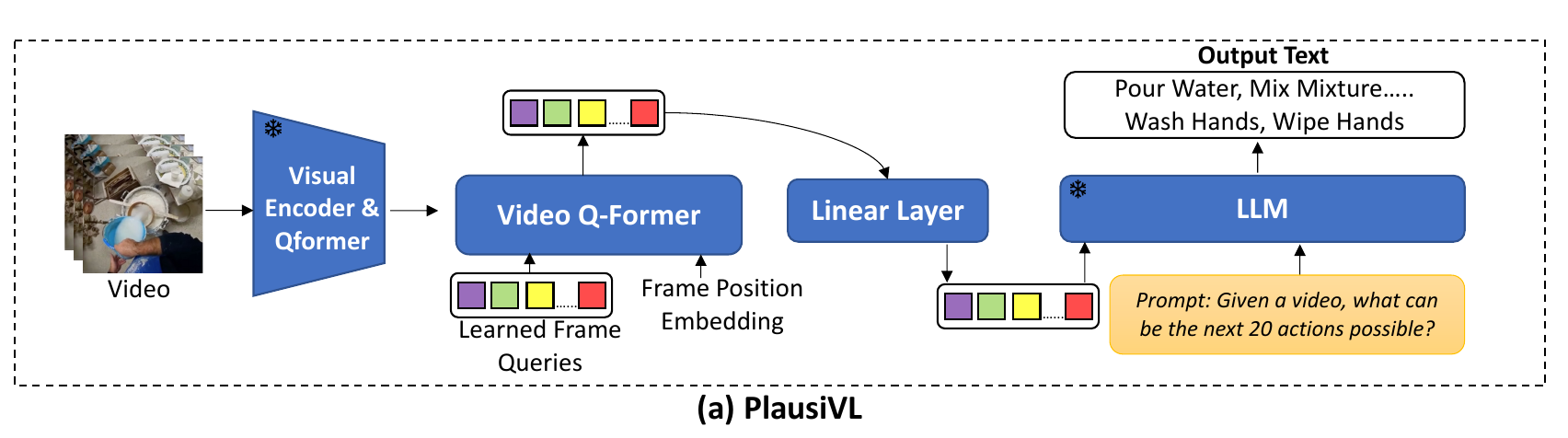}
    \includegraphics[trim=3.7cm 0cm 1.2cm 0.1cm, clip, width=0.9\linewidth]{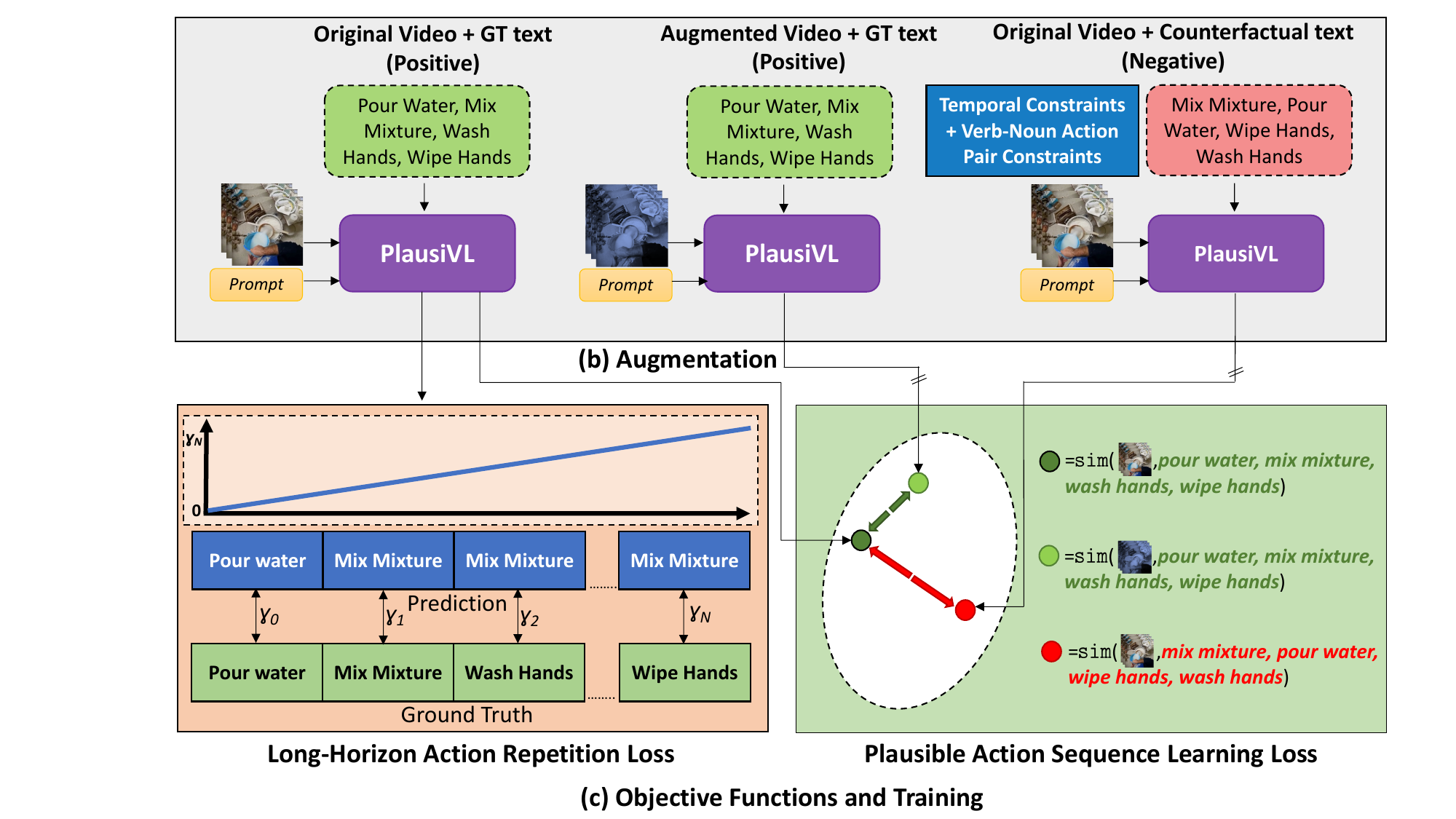}
    \vspace{-3mm}
    \caption{\textbf{Model diagram:(a) \method}: Given a video, a frozen visual encoder a Q-former with $k$ number of query tokens is used to extract frame level representations which are further concatenated with a frame position embedding layer to add temporal understanding. Next, the representations are passed through the video Q-former and a linear layer is added to project these features into the LLM space. These visual embeddings~(visual prompts) and are concatenated with text-prompts to get the desired output text~(Sec~\ref{sec:model_architecture}), \textbf{(b) Augmentation:} For plausible action anticipation, we use logical rules to create counterfactual implausible action sequences. Given an input video, we create a positive augmentation of the video and a negative augmentation by using temporal logical and verb-noun action pair constraints~(Sec~\ref{sec:Lplau}). \textbf{(c) Objective Functions and Training:} We train our model with two losses: (i) Plausible Action Sequence Learning Loss~($\Lplau$) which aligns the original video-plausible text pair closer to the positive augmentation of video-plausible text, and brings the original video-plausible text far apart from the video-counterfactual text.~(Sec~\ref{sec:Lplau}), (ii) long-horizon action repetition loss that ensures diverse and less repetitive actions by adding a higher penalty to the later tokens~(mix mixture and wipe hands) and lower penalty to immediate future actions~(pour water, pour water). The graph shows the linearly increasing $\gamma$ penalty for the tokens over the long-horizon~(Sec~\ref{sec:Lrep}).}
    \label{fig:model_diagram}
\end{figure*}

In the following sections, we present the details of our method, \method, to learn the temporal cues for plausible action sequence generation. 

\subsection{Model Architecture}
\label{sec:model_architecture}
Given a video clip of $N$ frames, $V = [v_{1}, v_{2}, v_{3}....v_{N}]$, we use a frozen visual encoder~(ViT) to extract video-frame-level representations, $V = [v'_{1}, v'_{2}, v'_{3}....v'_{N}]$. After this, each frame feature is passed through a Q-former~\cite{li2023blip} with $k$ number of query tokens, to get the $d_q$-dimensional visual representation as $v''_{i}\in\mathbb{R}^{k \times d_{q}}$. These queries are helpful in extracting the visual features with the most information aligned to the text. For the frames to have an understanding of the temporal relations among them, a frame position embedding layer is applied to each Q-former feature. At the same time, we also apply a clip-position embedding layer to infuse more grouping information about the frames that belong to a clip. These features are then passed through a video Q-former to aggregate the spatio-temporal information of the video. Finally, a linear projection layer is used to project these output representations to the LLM text embedding space of $d_l$ dimension, $v_{i}\in\mathbb{R}^{k_{l} \times d_{l}}$. These video embeddings can be considered as \textit{visual prompts} which are concatenated with the input text embeddings $t_i$ to make the LLM generate text conditioned on the video content. 


\section{Training}
\label{sec:training}
While the above backbone network ensures the alignment of the visual features with the LLM textual space, we also focus on making the model learn to better understand long-horizon temporal dependencies among the actions which is crucial for plausible action anticipation. To develop such temporal understanding in a model, we train our system to optimize two losses, (1). Plausible Action Sequence Learning loss $\Lplau$ and (2). Long-horizon action repetition loss $\Lrep$. With these two losses, the model can understand the temporal cues better to be able to generate a \textit{plausible} and diverse sequence of future actions.

\subsection{Plausible Action Sequence Learning loss}
\label{sec:Lplau}
For a model to be able to understand the plausible nature of an action sequence, it should be able to leverage the implicit temporal information present in input videos. Thus, we design a self-supervised plausible action sequence learning loss, $\Lplau$. The key idea is to create counterfactuals based on temporal logical constraints as well as verb-noun action pair logical constraints and optimize the network by minimizing a loss with two negative log-likelihood terms: (1) \textit{increase} the probability of associating the visual modality with the temporally correct and plausible sequence of actions, and (2) \textit{decrease} the probability of associating the video with the action sequences that are not plausible in the real-world and temporally incorrect. Here, sequences of action that satisfy the temporal as well as verb-noun action pair logic constraints are considered as logically correct.

\noindent \textbf{Temporal logical constraints}: In our work, we define a temporal constraint for an action sequence as follows: \textit{an action X that has to happen before an action Y} to make it a plausible sequence in the real-world. \noindent Consider for example, given a sequence of \textit{take eggs} $\rightarrow$ \textit{crack eggs} $\rightarrow$ \textit{whisk eggs} $\rightarrow$ \textit{cook omelette}, a counterfactual of this sequence of actions would be, \textit{take eggs} $\rightarrow$ \textit{cook omelette} $\rightarrow$ \textit{whisk eggs} $\rightarrow$ \textit{crack eggs} since \textit{crack eggs} would always happen before \textit{cook omelette}. Mathematically, we define it as follows:
\begin{align}
\small
    CF^{temp}(a_i, a_j) = \begin{dcases*}
        1, & if $ \forall_{t \in T} (t_{a_i} \rightarrow t_{a_j}) \wedge \neg (t_{a_j} \rightarrow t_{a_i}),  $ \\
        -1 & if $ \forall_{t \in T} (t_{a_j} \rightarrow t_{a_i}) \wedge \neg (t_{a_i} \rightarrow t_{a_j}),  $ \\
        0, & otherwise. 
        \end{dcases*}
\end{align}
\noindent where $CF^{temp}(a_i, a_j)$ is an action pair matrix with a value of 1 if $a_i$ always happens before $a_j$ for all the ground truth sequences $t \in T$, a value of -1 if $a_i$ always happens after $a_j$, and 0 otherwise if there is no relation between the two actions. With this temporal logical constraint, given a text sequence $t$, we perform a swap operation if there is a forward or backward relation between an action pair. Hence, given a ground truth text sequence $t$, we define the operation if $a_j$ always happens before $a_p$ as follows:
\begin{align}
\small
    t^{cf}(a_i, a_j, a_p, a_n) = \begin{dcases*}
        a_i, a_p, a_j, a_n, & if $ CF^{temp}(a_j, a_p)=1,  $ \\
        a_i, a_j, a_p, a_n, & otherwise. 
        \end{dcases*}
\end{align}
\noindent Similarly, we define the operation if $a_j$ always happens after $a_i$ as follows:
\begin{align}
\footnotesize
    t^{cf}(a_i, a_j, a_p, a_n) = \begin{dcases*}
        a_j, a_i, a_p, a_n, & if $CF^{temp}(a_j, a_i)=-1,$ \\
        a_i, a_j, a_p, a_n, & otherwise. 
        \end{dcases*}
\end{align}
\noindent Next, we define the another logical constraint - verb-noun action pair constraint. 

\noindent \textbf{Verb-Noun Action pair constraints}: For this, we create a counterfactual where a verb-noun action pair is not plausible in the real-world, for example, \textit{cook spoon}. We define a verb-noun action constraint as follows: \textit{a verb-noun pair consisting of an action verb that is plausible with the object noun} in the real-world. Mathematically, we define it as follows: 
\begin{align}
    CF^{act}(a_i, a_j) = \begin{dcases*}
        1, & if $ \forall_{t \in T} \neg ({a^v_i} \wedge {a^n_j}),  $ \\
        0, & otherwise. 
        \end{dcases*}
\end{align}
\noindent where $CF^{act}(a_i, a_j)$ is a verb-noun pair matrix with a value of 1 if for a verb, the corresponding noun is not plausible or vice-versa in all the ground truth actions $t \in T$ and 0 otherwise if the verb-noun pair is plausible. Similar to the temporal constraints mentioned above, with this verb-noun action pair constraint, given an action, we swap either the verb or noun with a uniform probability to create implausible verb-noun action pairs. Given a text action pair $t$, we define the operation of a counterfactual, implausible verb-noun action pair as follows:
\begin{dmath}
    t^{cf}(a^v_i, a^n_i) = \begin{dcases*}
        (a^v_i, a^n_j) || (a^v_j, a^n_i), & if $ CF^{act}(a^v_i, a^n_j) = 1,  $ \\
        (a^v_i, a^n_i), & otherwise. 
        \end{dcases*}
\end{dmath}
\noindent \textbf{Loss:} With this, for every video-text action sequence pair $(V_i, T_i)$ in the dataset $\mathcal{D}$, we create a temporal as well as verb-noun action pair counterfactual $T^{cf}_i$ for every textual ground truth text sequence and collect it as a dataset, $\mathcal{D}_{vtcf}$. Finally, we define plausible action sequence learning loss~($\Lplau$) as follows:
\begin{dmath}
	\Lplau = \mathbb{E}_{(v_i, t_i) \in \mathcal{D}_{\mathtt{vtcf}}}\biggl[
	- \log \Bigl(z(v_i, t_i, v'_i)\Bigr)
	- \log \left(1 - z(v_i, t_i, t^{cf}_i)\right)\biggr]
	\label{eq:Lplau}
\end{dmath}
\noindent In the above equation, $z(v_i, t_i, v'_i)$ and $z(v_i, t_i, t^{cf}_i)$ probabilities are computed as follows:
\begin{align}
    \centering
    z(v_i, t_i, v'_i) &= \sigma\Bigl(\mathtt{sim}( \Delta p(v_i, t_i), \Delta p(v'_i, t_i) )/ \tau\Bigr) \\
    z(v_i, t_i, t^{cf}_i) &= \sigma\left(\mathtt{sim}( \Delta p(v_i, t_i), \Delta p(v_i, t^{cf}_i) )/ \tau\right)
\end{align}
\noindent where $v_i$ and $v'_i$ are the visual embeddings of the original video and augmented video~(respectively), $t_i$ and $t^{cf}_i$ are the text embeddings of the ground truth text sequence and counterfactual text~(respectively), $\tau$ is the temperature, $\sigma$ is the sigmoid function, $\Delta p(., .)$ is the cross-modal video-text representation from LLM after passing through a MLP projection layer~(absorbed in the equation for better readability), and $\mathtt{sim}$ is the similarity function. \newline

\noindent In summary, training the model to optimize the $\Lplau$ loss helps the model to differentiate between the plausible and counterfactual/implausible action sequences by aligning the visual modality closer to the temporally correct, plausible action sequence. By learning this alignment, it is able to understand the implicit temporal information that defines the dependencies and correlations among actions in a plausible sequence.




\subsection{Long-Horizon Action Repetition Loss}
\label{sec:Lrep}
While the plausible action sequence learning loss $\Lplau$ helps the model to understand the implicit temporal information present in the action sequences, we consider another aspect of plausibility by reducing the repetition of actions and in turn generating more diverse actions. We observe that while the model is able to generate accurate, temporally correct, and diverse actions over a short temporal window, it starts repeating the same actions over a longer horizon. To mitigate this, we train the model by enforcing a larger penalty on the actions that happen over a longer horizon in the temporal window and lesser penalty to the actions that are immediately near to the observed video. We add a penalty of $\gamma_t$ over the negative log-likelihood of the probability as follows:
\begin{align}
    \centering
    p_t &= \frac{\exp(\hat{y}_t)}{\Sigma_j \exp(\hat{y}_j)}, \\
    \Lrep(p_t) &= -{\gamma_t}log(p_t)
    \label{eqn:Lrep}
\end{align}
\noindent where $\hat y_t$ is the output from the language model for the $t$'th token over which softmax operation is applied to get the probability $p_t$. $\gamma_t$ is the $\gamma$ value temporally unique to the $t$'th token following the order, ${\gamma_0} < {\gamma_1} < {\gamma_2} < \cdots < {\gamma_N}$.

\noindent In summary, by optimizing the $\Lrep$ loss, the model is penalized more for the actions that happen over a longer horizon and less penalized for immediate actions. This is helpful in regulating repetition and ensuring more diverse actions in the generated text. 

\noindent Finally, we train our model with the overall loss as:
\begin{equation}
    \centering
    \mathcal{L} = \alpha\Lplau + \beta\Lrep
    \label{eqn:bothlosses}
\end{equation}
\noindent where $\alpha$ and $\beta$ are the weight hyper-parameter for the two losses. 


\section{Experiments}
\label{sec:experiments}

\subsection{Implementation Details}
We process the videos of size $224 \times 224$ with Ego4D containing 8 clips with 4 frames, making a total of 32 frames, and \epickitchens with 32 frames as well. We use the pretrained Qformer model, BLIP2-FlanT5xxl from BLIP2~\cite{li2023blip} with number of query tokens as 32 and ViT-G/14 as our vision encoder. We train our method end-to-end with a learning rate of $1e^{-5}$, for 100 epochs, and $\alpha = 0.5$ and $\beta = 0.5$. We use LLaMA-2-7B as our language model. For long-horizon action repetition loss, $\Lrep$, we use $\gamma$ in the uniform distribution from $[0, 2]$ with number of steps equal to the number of output tokens from the language model. For plausible action sequence learning loss $\Lplau$, we use video augmentation of color jitter, random horizontal flip, and a random rotation of 10 degrees.

\subsection{Experimental Setup}

\noindent \textbf{Datasets:} We evaluate on two action anticipation datasets: Ego4D~\cite{grauman2022ego4d} and \epickitchens~\cite{damen2020rescaling}. Ego4D is a large-scale egocentric dataset covering diverse indoor and outdoor scenarios like home, workplace, etc. It consists of 3670 hours of videos with 115 verbs and 478 nouns. To evaluate our method on Ego4D, we use videos from the Forecasting and Hand-Object interaction subset and show results on the validation set. In Ego4D, a video and a stopping time is given, and the model predicts $N$ sets of sequences having $Z$ number of actions in the form of verb-noun pairs, $\{\{(\hat{v}_{z, n}, \hat{n}_{z, n})\}^{Z}_{z=1}\}_{n=1}^N$, where, $\hat{v}_{z, n}$ is the predicted verb and $\hat{n}_{z, n}$ is the predicted noun. 

\noindent \epickitchens~\cite{damen2020rescaling} is an egocentric dataset of a kitchen-based environment. It consists of 100 hours of egocentric videos with 97 verbs and 300 nouns. For this dataset, given an action segment that starts at time $\tau_{s}$, the model has to predict the anticipated action by observing a video segment of duration $[\tau_{s} - (\tau_{o} + \tau_{a}), \tau_{s} - \tau_{a}]$ where $\tau_{o}$ is the observation time and $\tau_{a}$ is the anticipation time. The anticipation time $\tau_{a}$ means how much in advance the model has to anticipate the action. 

\noindent \textbf{Baselines:} We compare our method with \visllms, Video-LLaMA~\cite{zhang2023video} and Video-LLM~\cite{chen2023videollm}. We also compare our method with the transformer and LSTM-based approaches for action anticipation along with text-based large language models for a more exhaustive analysis.

\noindent \textbf{Ablation Study:} In the ablation study, we present results of \method with and without $\Lplau$ and $\Lrep$ objective functions to show the effect of each component on the final performance of the model.

\begin{table}[]
\centering
\begin{tabular}{lcc} \hline
\multicolumn{1}{c||}{\multirow{2}{*}{\textbf{Method}}} & \multicolumn{2}{c}{\textbf{ED@(Z=20)} $\downarrow$}                    \\ \cline{2-3} 
\multicolumn{1}{c||}{} & \multicolumn{1}{c}{\textbf{Verb}} & \multicolumn{1}{c}{\textbf{Noun}}  \\ \hline
\multicolumn{1}{l||}{RepLAI~\cite{mittal2022learning}} & 0.755                      & \multicolumn{1}{l}{0.834} \\ 
\multicolumn{1}{l||}{SlowFast~\cite{grauman2022ego4d}} & 0.745                      & \multicolumn{1}{l}{0.779} \\ 
\multicolumn{1}{l||}{ICVAE~\cite{mascaro2022intention}} & 0.741                      & \multicolumn{1}{l}{0.739} \\ 
\multicolumn{1}{l||}{HierVL~\cite{ashutosh2023hiervl}} & 0.723                      & \multicolumn{1}{l}{0.734} \\ 
\multicolumn{1}{l||}{Video+CLIP~\cite{das2022video+}} & 0.715                      & \multicolumn{1}{l}{0.748} \\ \hline
\multicolumn{1}{l||}{{AntGPT~\cite{zhao2023antgpt}}} & 0.700                      & \multicolumn{1}{l}{0.717} \\ 
\multicolumn{1}{l||}{Video LLM~\cite{chen2023videollm}} & 0.721                    & \multicolumn{1}{l}{0.725}  \\ 
\multicolumn{1}{l||}{Video LLaMA~\cite{zhang2023video}} & \multicolumn{1}{l}{0.703} & \multicolumn{1}{l}{0.721}  \\ \hline
\multicolumn{1}{l||}{\textbf{\method}}  & \multicolumn{1}{l}{\textbf{0.679}}  & \multicolumn{1}{l}{\textbf{0.681}} \\ \hline
\end{tabular}
\caption{Performance on Long-term action anticipation on Ego4D $\downarrow$: Lower is better. This shows shows that our method, \method is able to outperform all the previous baselines for verb, noun, and action.} 
\label{tbl:mainego4d}
\end{table}
\begin{table}[]
\centering
\begin{tabular}{lccccc}\hline
\multicolumn{1}{c||}{\multirow{2}{*}{\textbf{Method}}} & \multicolumn{3}{c}{\begin{tabular}[c]{@{}c@{}}\textbf{Class-mean} \\ \textbf{Top-5 recall (\%)} $\uparrow$ \end{tabular}}                                 \\  \cline{2-4}
\multicolumn{1}{c||}{} & \multicolumn{1}{c}{\textbf{Verb}} & \multicolumn{1}{c}{\textbf{Noun}} & \multicolumn{1}{c}{\textbf{Action}} \\ \hline

\multicolumn{1}{l||}{RU-LSTM~\cite{damen2020rescaling}} & \multicolumn{1}{c}{23.20}                        & \multicolumn{1}{c}{31.40}                        & \multicolumn{1}{c}{14.70}  \\
\multicolumn{1}{l||}{Temporal Aggregation~\cite{sener2020temporal}} & \multicolumn{1}{c}{27.80} & \multicolumn{1}{c}{30.80} & \multicolumn{1}{c}{14.00}  \\
\multicolumn{1}{l||}{Video LLM~\cite{chen2023videollm}} & \multicolumn{1}{c}{-}                        & \multicolumn{1}{c}{-}                        & \multicolumn{1}{c}{15.40}  \\ 
\multicolumn{1}{l||}{AFFT~\cite{zhong2023anticipative}} & \multicolumn{1}{c}{22.80}                        & \multicolumn{1}{c}{34.60}                        & \multicolumn{1}{c}{18.50}  \\
\multicolumn{1}{l||}{AVT~\cite{girdhar2021anticipative}} & \multicolumn{1}{c}{28.20}                        & \multicolumn{1}{c}{32.00}                        & \multicolumn{1}{c}{15.90}  \\
\multicolumn{1}{l||}{MeMViT~\cite{wu2022memvit}} & \multicolumn{1}{c}{32.20}                        & \multicolumn{1}{c}{37.00}                        & \multicolumn{1}{c}{17.70}  \\ 
\multicolumn{1}{l||}{RAFTformer~\cite{girase2023latency}} & \multicolumn{1}{c}{33.80}                        & \multicolumn{1}{c}{37.90}                        & \multicolumn{1}{c}{19.10} \\ 
\multicolumn{1}{l||}{InAViT~\cite{roy2022interaction}} & \multicolumn{1}{c}{52.54}                        & \multicolumn{1}{c}{51.93}                        & \multicolumn{1}{c}{25.89}  \\ 
\multicolumn{1}{l||}{Video LLaMA~\cite{zhang2023video}} &  \multicolumn{1}{c}{52.90}                        &    \multicolumn{1}{c}{52.01}                      & \multicolumn{1}{c}{26.05}       \\ \hline
\multicolumn{1}{l||}{\textbf{\method}} & \multicolumn{1}{c}{\textbf{55.62}}   &  \multicolumn{1}{c}{\textbf{54.23}}   & \multicolumn{1}{c}{\textbf{27.60}}       \\ \hline
\end{tabular}
\caption{Performance of action anticipation on \epickitchens on class-mean Top-5 recall~(\%) $\uparrow$): Higher is better. Our method is able to outperform all the previous baselines.}
\label{tbl:mainek100}
\vspace{-2mm}
\end{table}
\begin{table}[]
\small
\begin{tabular}{cclllll}
                                            &                                            & \multicolumn{2}{c}{\textbf{Ego4D}}                              & \multicolumn{3}{c}{\textbf{EPIC-Kitchens-100}}                                                        \\ \hline
\multicolumn{1}{c}{\multirow{2}{*}{$\Lplau$}} & \multicolumn{1}{c|}{\multirow{2}{*}{$\Lrep$}} & \multicolumn{2}{c|}{\textbf{ED@(Z=20)} $\downarrow$}                         & \multicolumn{3}{c}{\begin{tabular}[c]{@{}c@{}}\textbf{Class-mean} \\ \textbf{Top-5 recall (\%)} $\uparrow$ \end{tabular}} \\ \cline{3-7} 
\multicolumn{1}{c}{}                       & \multicolumn{1}{c|}{}                      & \multicolumn{1}{c}{\textbf{Verb}} & \multicolumn{1}{c|}{\textbf{Noun}}   & \multicolumn{1}{c}{\textbf{Verb}}     & \multicolumn{1}{c}{\textbf{Noun}}     & \multicolumn{1}{c}{\textbf{Action}}    \\ \hline
\multicolumn{1}{c}{$\checkmark$}   & \multicolumn{1}{c|}{$\checkmark$}   &      \multicolumn{1}{l}{0.679}               & \multicolumn{1}{l|}{0.683}  &    \multicolumn{1}{l}{55.62}                         &     \multicolumn{1}{l}{54.23}                        & \multicolumn{1}{l}{27.60} \\
\multicolumn{1}{c}{$\checkmark$} & \multicolumn{1}{c|}{}  &   \multicolumn{1}{l}{0.686}   & \multicolumn{1}{l|}{0.698} & \multicolumn{1}{l}{54.50}  & \multicolumn{1}{l}{53.60}  & \multicolumn{1}{l}{26.67} \\
\multicolumn{1}{c}{}  & \multicolumn{1}{c|}{$\checkmark$} & \multicolumn{1}{l}{0.691}  & \multicolumn{1}{l|}{0.707}  &      \multicolumn{1}{l}{54.15}                       &      \multicolumn{1}{l}{53.03}                       & \multicolumn{1}{l}{26.21}         \\  
\multicolumn{1}{c}{}  & \multicolumn{1}{c|}{} & \multicolumn{1}{l}{0.703}  & \multicolumn{1}{l|}{0.721}  &      \multicolumn{1}{l}{52.90}                       &      \multicolumn{1}{l}{52.01}                       & \multicolumn{1}{l}{26.05}         \\  \hline
\end{tabular}
\caption{Ablation study of modules, $\Lplau$ and $\Lrep$ in our method on Ego4D $\downarrow$: Lower is better, and \epickitchens on class-mean Top-5 recall~(\%) $\uparrow$): Higher is better. The analysis that starting from our method, row (1), there is a dip in the performance as each module is removed showing that the losses, $\Lplau$ and $\Lrep$ are helpful in improving the performance.}
\vspace{-2mm}
\label{tbl:ablationego4dek100}
\end{table}

\begin{table}[]
\small
\begin{tabular}{@{}clllll@{}}
                                                                                        & \multicolumn{2}{c}{\textbf{Ego4D}}                              & \multicolumn{3}{c}{\textbf{EPIC-Kitchens-100}}                                                        \\ \hline
\multicolumn{1}{@{}c|}{\multirow{2}{*}{\textbf{Method}}}  & \multicolumn{2}{c|}{\textbf{ED@(Z=20)} $\downarrow$}                         & \multicolumn{3}{c}{\begin{tabular}[c]{@{}c@{}}\textbf{Class-mean} \\ \textbf{Top-5 recall (\%)} $\uparrow$ \end{tabular}} \\ \cline{2-6} 
 \multicolumn{1}{c|}{}                      & \multicolumn{1}{c}{\textbf{Verb}} & \multicolumn{1}{c|}{\textbf{Noun}}   & \multicolumn{1}{c}{\textbf{Verb}}     & \multicolumn{1}{c}{\textbf{Noun}}     & \multicolumn{1}{@{}c@{}}{\textbf{Action}}    \\ \hline
\multicolumn{1}{@{}c@{}|}{PlausiVL (w/ DNR)}  &      \multicolumn{1}{l}{0.689}               & \multicolumn{1}{l|}{0.695}  &    \multicolumn{1}{l}{54.30}                         &     \multicolumn{1}{l}{53.20}                        & \multicolumn{1}{l}{26.63} \\
\multicolumn{1}{@{}c@{}|}{PlausiVL}  &   \multicolumn{1}{l}{0.679}   & \multicolumn{1}{l|}{0.681} & \multicolumn{1}{l}{55.62}  & \multicolumn{1}{l}{54.23}  & \multicolumn{1}{l}{27.60} \\ \hline
\end{tabular}
\caption{Performance of \method with and without ``DNR: Do NOT repeat actions" in the prompt. We can observe that having DNR in the prompt does not give much improvement in the performance as compared to training the model with long-horizon action repetition loss~($\Lrep$) as objective function.}
\vspace{-2mm}
\label{tbl:dnr_big}
\end{table}


\begin{table}[]
\small
\centering
\begin{tabular}{l|c|c}
\hline
                     & \textbf{BLEU Score $\uparrow$} & \textbf{Repetition Score $\downarrow$} \\ \hline
                 
Video-LLaMA~\cite{zhang2023video} &      37.89                      &   7.12                 \\
\textbf{\method}        &      \textbf{45.54}                     &    \textbf{5.87}                 \\ \hline
\textcolor{gray}{Ground Truth}        &       \textcolor{gray}{100.00}   & \textcolor{gray}{4.33}     \\  \hline
\end{tabular}
\caption{BLEU score and Repetition Score on the Ego4D dataset. For BLEU score, $\uparrow$: Higher is better, and for repetition score, $\downarrow$: lower is better. We can observe that both the BLEU score and repetition score are better for \method than Video-LLaMA.}
\label{tbl:bleurep}
\vspace{-2mm}
\end{table}
\begin{table*}[]
\centering
\begin{tabular}{lcccccccccc}\hline
\multicolumn{1}{c||}{\multirow{2}{*}{\textbf{Method}}} & \multicolumn{3}{c|}{\begin{tabular}[c]{@{}c@{}}\textbf{Unseen} $\uparrow$ \end{tabular}} & \multicolumn{3}{c}{\begin{tabular}[c]{@{}c@{}}\textbf{Tail} $\uparrow$ \end{tabular}}                                 \\  \cline{2-7}
\multicolumn{1}{c||}{} & \multicolumn{1}{c}{\textbf{Verb}} & \multicolumn{1}{c}{\textbf{Noun}} & \multicolumn{1}{c|}{\textbf{Action}} & \multicolumn{1}{c}{\textbf{Verb}} & \multicolumn{1}{c}{\textbf{Noun}} & \multicolumn{1}{c}{\textbf{Action}} \\ \hline

\multicolumn{1}{l||}{RU-LSTM~\cite{damen2020rescaling}} & \multicolumn{1}{c}{28.78}                        & \multicolumn{1}{c}{27.22}                        & \multicolumn{1}{c|}{14.15} & \multicolumn{1}{c}{19.77}                        & \multicolumn{1}{c}{22.02}                        & \multicolumn{1}{c}{11.14}  \\
\multicolumn{1}{l||}{Temporal Aggregation~\cite{sener2020temporal}} & \multicolumn{1}{c}{28.80} & \multicolumn{1}{c}{27.20} & \multicolumn{1}{c|}{14.20} & \multicolumn{1}{c}{19.80} & \multicolumn{1}{c}{22.00} & \multicolumn{1}{c}{11.10}  \\
\multicolumn{1}{l||}{Video LLM~\cite{chen2023videollm}} & \multicolumn{1}{c}{-}                        & \multicolumn{1}{c}{-}                        & \multicolumn{1}{c|}{12.60} & \multicolumn{1}{c}{-}                        & \multicolumn{1}{c}{-}                        & \multicolumn{1}{c}{12.00}  \\ 
\multicolumn{1}{l||}{AFFT~\cite{zhong2023anticipative}} & \multicolumn{1}{c}{24.80}                        & \multicolumn{1}{c}{26.40}                        & \multicolumn{1}{c|}{15.50}  & \multicolumn{1}{c}{15.00}                        & \multicolumn{1}{c}{27.70}                        & \multicolumn{1}{c}{16.20}   \\
\multicolumn{1}{l||}{AVT~\cite{girdhar2021anticipative}} & \multicolumn{1}{c}{29.50}                        & \multicolumn{1}{c}{23.90}                        & \multicolumn{1}{c|}{11.90} & \multicolumn{1}{c}{21.10}                        & \multicolumn{1}{c}{25.80}                        & \multicolumn{1}{c}{14.10}  \\
\multicolumn{1}{l||}{MeMViT~\cite{wu2022memvit}} & \multicolumn{1}{c}{28.60}                        & \multicolumn{1}{c}{27.40}                        & \multicolumn{1}{c|}{15.20} & \multicolumn{1}{c}{25.30}                        & \multicolumn{1}{c}{31.00}                        & \multicolumn{1}{c}{15.50}  \\ 
\multicolumn{1}{l||}{InAViT~\cite{roy2022interaction}} & \multicolumn{1}{c}{46.45}                        & \multicolumn{1}{c}{51.30}                        & \multicolumn{1}{c|}{25.33}  & \multicolumn{1}{c}{45.34}                        & \multicolumn{1}{c}{39.21}                        & \multicolumn{1}{c}{20.22}  \\ 
\multicolumn{1}{l||}{Video LLaMA~\cite{zhang2023video}} &  \multicolumn{1}{c}{46.87}                        &    \multicolumn{1}{c}{51.47}                      & \multicolumn{1}{c|}{25.40}  &  \multicolumn{1}{c}{45.71}                        &    \multicolumn{1}{c}{39.32}                      & \multicolumn{1}{c}{20.35}       \\ \hline
\multicolumn{1}{l||}{\textbf{\method}} & \multicolumn{1}{c}{\textbf{49.50}}   &  \multicolumn{1}{c}{\textbf{53.90}}   & \multicolumn{1}{c|}{\textbf{27.01}} & \multicolumn{1}{c}{\textbf{48.44}}   &  \multicolumn{1}{c}{\textbf{41.29}}   & \multicolumn{1}{c}{\textbf{22.10}}       \\ \hline
\end{tabular}
\caption{Performance of action anticipation on \epickitchens Unseen Participants and Tail Classes on class-mean Top-5 recall~(\%) $\uparrow$): Higher is better. Our method is able to outperform all the previous baselines.}
\label{tbl:unseentailek100}
\vspace{-2mm}
\end{table*}

\subsection{Discussion of Results}
Referring to Table~\ref{tbl:mainego4d} and Table~\ref{tbl:mainek100}, we can observe that \method is able to perform better when compared with the baselines. This can be attributed to its ability to be able to understand the plausibility in the action sequences and leverage the temporal correlations among the actions in a sequence. We present a closer analysis of the results in our discussion following next.

\begin{figure*}[t]
    \centering
    \includegraphics[trim=0cm 4.5cm 0cm 0cm, clip, width=0.99\linewidth]{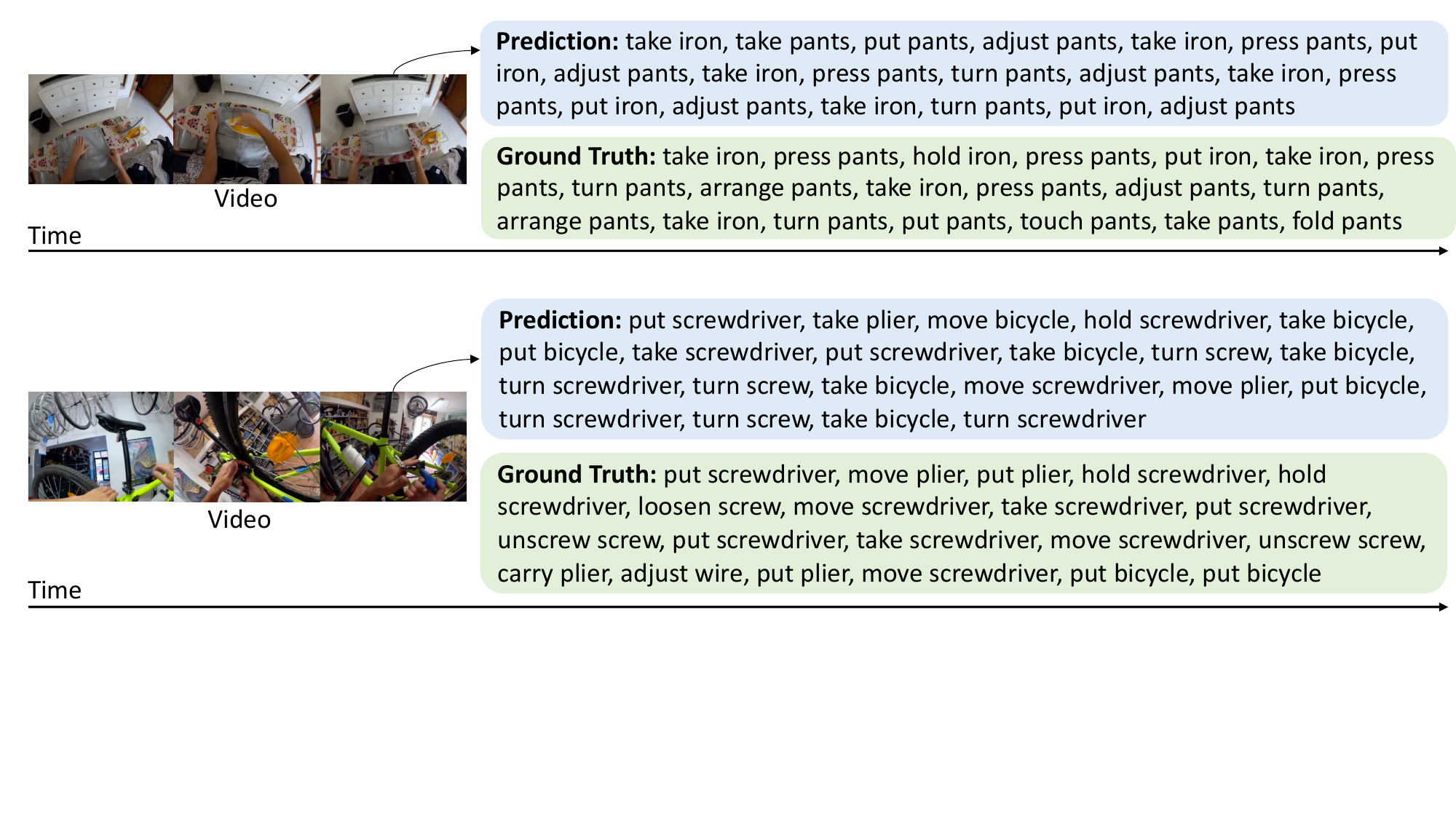}
    \vspace{-2mm}
    \caption{Qualitative Results: Given a video, the top \textcolor{blue}{blue} box shows the prediction from \method and the \textcolor{Green}{green} box contains the ground truth action sequence for reference. We can observe that \method is able to generate action sequences that satisfy the temporal logic constraints and are diverse with less repetitions. The predicted action sequence is also closer to the ground truth action sequence.}
    \label{fig:qual_results}
    \vspace{-2mm}
\end{figure*}

\noindent \textbf{\method shows performance gain towards action anticipation:} Prior \visllms~\cite{zhang2023video,chen2023videollm} have only explored the visual-text alignment and lack the temporal understanding needed for the action anticipation. To show that our model is able to learn the temporal understanding, we compare \method with Video-LLM and Video-LLaMA in Table~\ref{tbl:mainego4d} and observe an improvement of 4.2\% and 2.4\%, respectively on verbs. Similarly, we observe an improvement of 2.72\% and 2.22\% on verbs for \epickitchens in Table~\ref{tbl:mainek100}. The improvement in the performance emphasizes that the model is able to learn the temporal dependencies among the actions to generate more accurate and plausible action sequences. Qualitative results presented in Figure~\ref{fig:qual_results} also show the quality of our generated sequence in comparison to the ground truth. We can see that our method is able to understand the activity happening the video and anticipate the temporal future action sequence accordingly. We also exhaustively compare \method with prior approaches in Table~\ref{tbl:mainego4d} and Table~\ref{tbl:mainek100} that utilize transformer and LSTM-based architectures and show that our method is able to perform better. 

\noindent \textbf{$\Lplau$ helps the model to learn plausible future action sequences:} We hypothesize that for generating accurate future action sequences, a model should have an understanding about the temporal plausibility of an action sequence in the real-world. To assess if our devised loss function, plausible action sequence learning loss, $\Lplau$ is able to create such understanding in the model, we compare our method, row (1) and our method without $\Lplau$, rows (3) and (4) in Table~\ref{tbl:ablationego4dek100}. We observe by removing this module, there is a drop in the performance of 1.2~\% on verbs for Ego4D and 1.47~\% for verbs of \epickitchens~(row(1) and row(3) are compared). This shows that training a model with $\Lplau$ as an objective function helps the model to learn the implicit temporal information of action correlations in a sequence. Through learning to differentiate between the plausible and not plausible action sequences and aligning the video representations closer to the plausible action sequences, the model learns an effective video-text alignment which helps in generating more accurate, plausible future action sequences.

\noindent \textbf{$\Lrep$ helps with lesser repetition and more diversity over long horizons:} We also try to address another aspect of plausibility in action sequences by making the model learn to generate sequences with less repetitive actions and more diverse actions via our devised objective function, long-horizon action repetition loss, $\Lrep$. To assess the efficacy of this module, we compare our method, row (1) and our method without $\Lrep$, row (2) and row (4) in Table~\ref{tbl:ablationego4dek100}. We observe that there is performance dip of 1.5~\% on Ego4D nouns and 0.63~\% on \epickitchens nouns. This indicates that by penalizing the actions more over the long horizon, $\Lrep$ is able to reduce the repetition of actions in the sequence generation and hence, contribute towards plausible action anticipation sequences.


\noindent \textbf{Training with $\Lrep$ loss vs prompt tuning:} We perform an analysis where instead of training the model with $\Lrep$ objective function, we simply prompt the model with the phrase: ``Do NOT repeat actions"~(DNR). We compare \method trained with $\Lplau$ and $\Lrep$ losses~(row 2) and \method trained with $\Lplau$ and DNR prompt~(row 1) and present the results of this analysis for Ego4D and \epickitchens in Table~\ref{tbl:dnr_big}. We can observe that simply prompting the model with DNR in the prompt does not give much improvement in the performance as compared to training the model with long-horizon action repetition loss~($\Lrep$) as objective function. Training the model $\Lrep$ penalizes the model for repeating the actions and makes the model learn to generate more diverse actions. This penalty is helpful in reducing repetition of the actions over a long-horizon. Simply stating DNR in the prompt only gives an instruction/command to the model, whereas, training the model with $\Lrep$ loss influences the learning of the model which is needed for the task of action anticipation. \newline

\noindent \textbf{\Visllm vs Text-large-language-model:} Given the exploration of text-only large language models, we also address the comparison between text-based LLM and \visllms for the task of action anticipation. We compare \method with AntGPT~\cite{zhao2023antgpt} which is a text-based LLM and observe a performance gain of 2.1\% on verbs and 3.6\% on nouns for Ego4D from our method. We reason that a major drawback of text-based LLM for this task is that they completely discard the visual as well as temporal information present in the videos. Whereas, the task of action anticipation is highly dependent on the visual spatio-temporal information to understand the real-world temporal flow of actions and anticipate actions accurately. Incorporating visual modality can give crucial information such as the environment of the agent, the objects interacted with, and other objects in the scene which might be interacted with later in the future. Such vital information is lost when converting a video into textual actions~\cite{zhao2023antgpt} or into a summary~\cite{huang2023palm}. Summarizing a video into text-based information can only provide the high-level details about a video, but it doesn't give a signal about the real-world temporal flow of the actions and objects in a video.


\noindent \textbf{\method is able to generate plausible action sequences:} To further emphasize the plausibility, less repetition and quality our generated text, we compute the BLEU score~\cite{papineni2002bleu} and repetition score. The repetition score is an average of the number of actions that are repeated in an action sequence and the BLEU score measures the similarity between our generated text and ground truth. We report the results in Table~\ref{tbl:bleurep}. By having a better BLEU score than the baseline, we show that the generated text from our method is a more plausible action sequence, thus emphasizing the efficacy of our objective functions. Similarly, by having a lower repetition score than the baseline, we show that the model has lesser repetitive actions in the generated sequence. Our method repeats 5.87 actions in an action sequence on average whereas Video-LLaMA repeats an average of 7.12 actions. We also observe an average repetition of 4.33 actions in ground truth action sequences. Moreover, a lower edit distance metric in Table~\ref{tbl:mainego4d} also indicates less repetition and more plausibility in the generated text as a lower metric would mean less substitutions were made to bring the output text closer to the ground truth.\newline

\noindent \textbf{Generalization and robustness to long-tail:} We evaluate our method on the unseen participants and tail classes of \epickitchens~\cite{damen2020rescaling} and present the results in Table~\ref{tbl:unseentailek100}. Unseen participants consists of those participants that are not present in the train set and tail classes are defined to be the smallest classes whose instances are around 20\% of the total number of instances in the train set. We observe that a better performance of our approach on the unseen participants as compared to the other baselines shows the generalizability of our model across unseen data. Similarly, a better performance on the tail classes shows that our model is robust to the long-tail distribution of the \epickitchens dataset. \newline

\begin{figure}[t]
    \centering
    \includegraphics[trim=0.2cm 0.2cm 0.3cm 0.3cm, clip, width=0.93\linewidth]{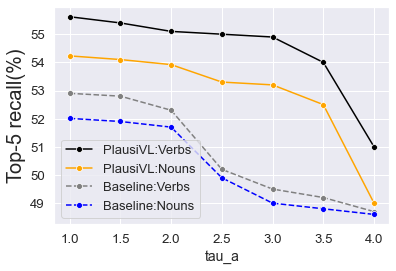}
    \vspace{-2mm}
    \caption{Analysis of $\tau_a$ vs. verb-noun class-mean Top-5 recall (\%) accuracy~($\uparrow$) on EK100.}
    \label{fig:taua_accuracy}
    \vspace{-2mm}
\end{figure}


\noindent \textbf{Anticipation time $\tau_a$ vs Accuracy:} 
$\tau_a$ is the anticipation time between the end time of observed video and the starting time of the first action to be anticipated. The video during the anticipation period $\tau_a$ is unobserved. For EK100, $\tau_a$=1s and for Ego4D, $\tau_a$=2.20s on an average. We analyze changing $\tau_a$ versus accuracy on EK100 in Figure~\ref{fig:taua_accuracy}. We can observe that the method is quite robust till $\tau_a$=3.5s whereas Video-LLaMA is only robust till $\tau_a$=2.0s for EK100. This shows that the model can predict future actions even with a far anticipation time.

\section{Conclusion}
\label{sec:conclusion}
In this work, we leverage the generative capabilities of \visllms for plausible action anticipation. In addition to the abilities of \visllms, for the model to better understand the plausibility in an action sequence, we introduce a plausible action sequence learning loss which helps the model to differentiate between the plausible and not plausible action sequences, and thus learn anticipation related temporal cues. We further devise a long-horizon action repetition loss that puts a higher penalty on the actions that happen over a longer temporal window and are more prone to repetition, thus mitigating action repetition and ensuring more diverse actions. Experimental results show that our model is able to perform better by generating more plausible and accurate action sequences on Ego4D and \epickitchens. While our method is an initial step towards plausible action anticipation, there can be further exploration mitigating the issue of hallucinating implausible action sequences in the future work. 

{
    \small
    \bibliographystyle{ieeenat_fullname}
    \bibliography{main}
}

\clearpage
\appendix

\section{Implementation Details}
We train our method end-to-end with a batch size of 2 for Ego4D and 4 for \epickitchens, linear warmup cosine as learning rate scheduler, along with the pre-trained weights of Video-LLaMA~\cite{zhang2023video} on 2 A6000 GPUs for 2.5 days.  

\subsection{Metrics}
\textbf{Edit Distance~(ED@(Z=20))~\cite{grauman2022ego4d}:} This metric is computed over a sequence of verb and noun predictions using the Damerau-Levenshtein distance~\cite{damerau1964technique,levenshtein1966binary} and takes into account the sequential nature of the action anticipation task. A prediction is considered correct if it matches the ground truth at a specific time step using the edit distance operations - insertion, deletion, substitution, and transposition. A total of $K$ predictions are evaluated and the smallest edit distance between a prediction and ground truth is reported~\cite{grauman2022ego4d}. We consider the value of $Z=20$ and $K=5$ which is the same as Ego4D~\cite{grauman2022ego4d}. \newline

\noindent \textbf{Class-mean Top-5 Recall~(\%)~\cite{damen2020rescaling}:} This metric evaluates if the ground truth class is within the top-5 predictions and averages the per-class performance to equally weight all the classes. The top-k criterion takes into account the uncertainty/multi-modality in the future action prediction and class-mean is helpful for balancing the long-tail distribution.

\begin{figure}[t]
    \centering
    \includegraphics[width=\linewidth]{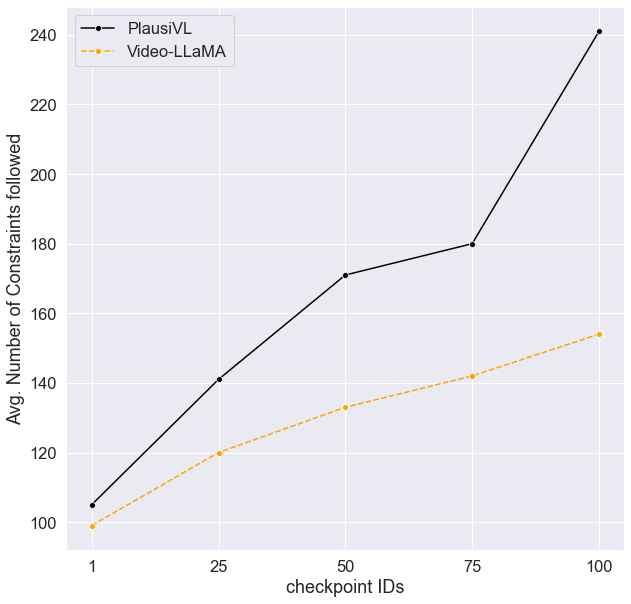}
    \vspace{-5mm}
    \caption{Analysis of plausibility in generated action sequence: \textcolor{black}{Black} line represents our method and \textcolor{orange}{orange} is the baseline, Video-LLaMA. Comparing the two line plots, we can observe that \method follows more number of temporal and action constraints over training than Video-LLaMA indicating that the objective functions $\Lplau$ and $\Lrep$ are helping the model to learn temporal cues needed to generate plausible action sequences for action anticipation.}
    \label{fig:constraint_graph}
\end{figure}

\section{Quantitative Analysis}




\noindent \textbf{Analysis of plausibility in generated action sequence:} To evaluate if the generated text is a plausible action sequence and additionally, the efficacy of the $\Lplau$ and $\Lrep$ objective functions, we calculate the average number of temporal and action constraints followed in the generated text. We compare the average number of constraints followed by \method versus the baseline Video-LLaMA~\cite{zhang2023video} and present the graph visualization in Figure~\ref{fig:constraint_graph}. We report the average number of constraints followed over the training and show the number over the checkpoints from beginning till the end of training. From the figure, we can observe that as the training of the model with $\Lplau$ and $\Lrep$ losses progresses, the average number of constraints followed increases in the generated text. Morever, the average number of \method is higher than that of Video-LLaMA. This indicates that by training the model with $\Lplau$ and $\Lrep$ objective functions, the model can generate more plausible action sequences and they help the model learn the implicit temporal information needed for plausible action anticipation. \newline


\begin{table}[]
\scriptsize
\centering
\begin{tabular}{l|cc|cc|cc}
\hline
\multicolumn{1}{c|}{\multirow{2}{*}{\textbf{Method}}} & \multicolumn{2}{c|}{\textbf{n\_rep=2}} & \multicolumn{2}{c|}{\textbf{n\_rep=3}} & \multicolumn{2}{c}{\textbf{n\_rep=4}} \\ \cline{2-7} 
\multicolumn{1}{c|}{}                                 & \textbf{Verb}      & \textbf{Noun}     & \textbf{Verb}      & \textbf{Noun}     & \textbf{Verb}     & \textbf{Noun}     \\ \hline
Video-LLaMA  &   0.703  &  0.721   &  0.704  &  0.724  &  0.704   &   0.726 \\
PLausiVL  &  0.680  &  0.681  &  0.679  &  0.681 & 0.680 & 0.683 \\ \hline
\end{tabular}
\vspace{-1em}
\caption{Results on different n\_rep for Ego4D on ED@(Z=20) $\downarrow$}
\label{tbl:suppreb_rep_analysis}
\end{table}
\begin{table}[]
\centering
\begin{tabular}{lcc} \hline
\multicolumn{1}{c||}{\textbf{Method}} & \multicolumn{1}{c}{\textbf{Verb}} & \multicolumn{1}{c}{\textbf{Noun}}  \\ \hline
\multicolumn{1}{l||}{CLR Paradigm} &        0.726          & \multicolumn{1}{l}{0.766} \\ \hline
\multicolumn{1}{l||}{\method w/ $\Lplau$} & 0.686 & \multicolumn{1}{l}{0.698} \\ 
\multicolumn{1}{l||}{\textbf{\method}}  & \multicolumn{1}{l}{\textbf{0.679}}  & \multicolumn{1}{l}{\textbf{0.681}} \\ \hline
\end{tabular}
\vspace{-1em}
\caption{Contrastive Loss with negative sample from other videos~(CLR Paradigm) for Ego4D on ED@(Z=20) $\downarrow$}
\label{tbl:r2_clr}
\end{table}

\noindent \textbf{$\Lrep$ loss is dataset independent}: We perform an analysis to highlight that repetition loss is independent of the dataset. In other words, the performance of the repetition loss does not depend on the number of repeated actions in a dataset. We present this analysis in Table~\ref{tbl:suppreb_rep_analysis}. We observe no strong correlation between n\_rep and performance, showing data-independency and also show that \method w/ repetition loss reduces repetition and outperforms the baseline. \newline

\noindent \textbf{Different videos as negative samples for $\Lplau$ loss:} For the $\Lplau$ loss, we use an implausible action sequence as a negative sample. We perform an analysis of using negative samples from other videos and show the results in Table~\ref{tbl:r2_clr}. This setting performs worse than Row 2,3 as it gives a weaker signal of counterfactual temporal plausibility than the signal of an implausible action sequence, since sequences from other videos are also temporally plausible.

\begin{figure*}[t]
    \centering
    \includegraphics[trim=0cm 4.5cm 0cm 0cm, clip, width=0.93\linewidth]{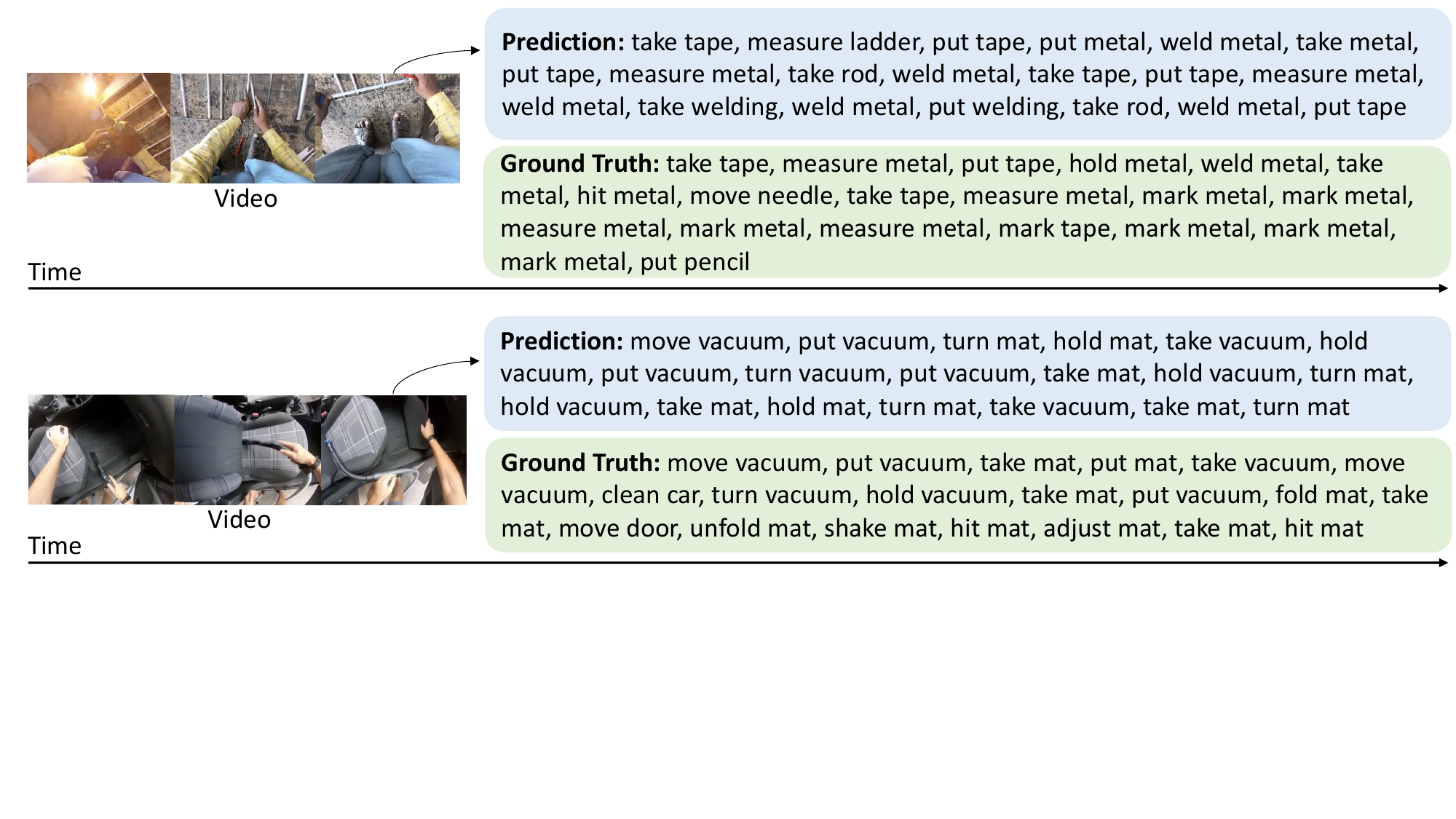}
    \includegraphics[trim=0cm 3.5cm 0cm 0cm, clip, width=0.93\linewidth]{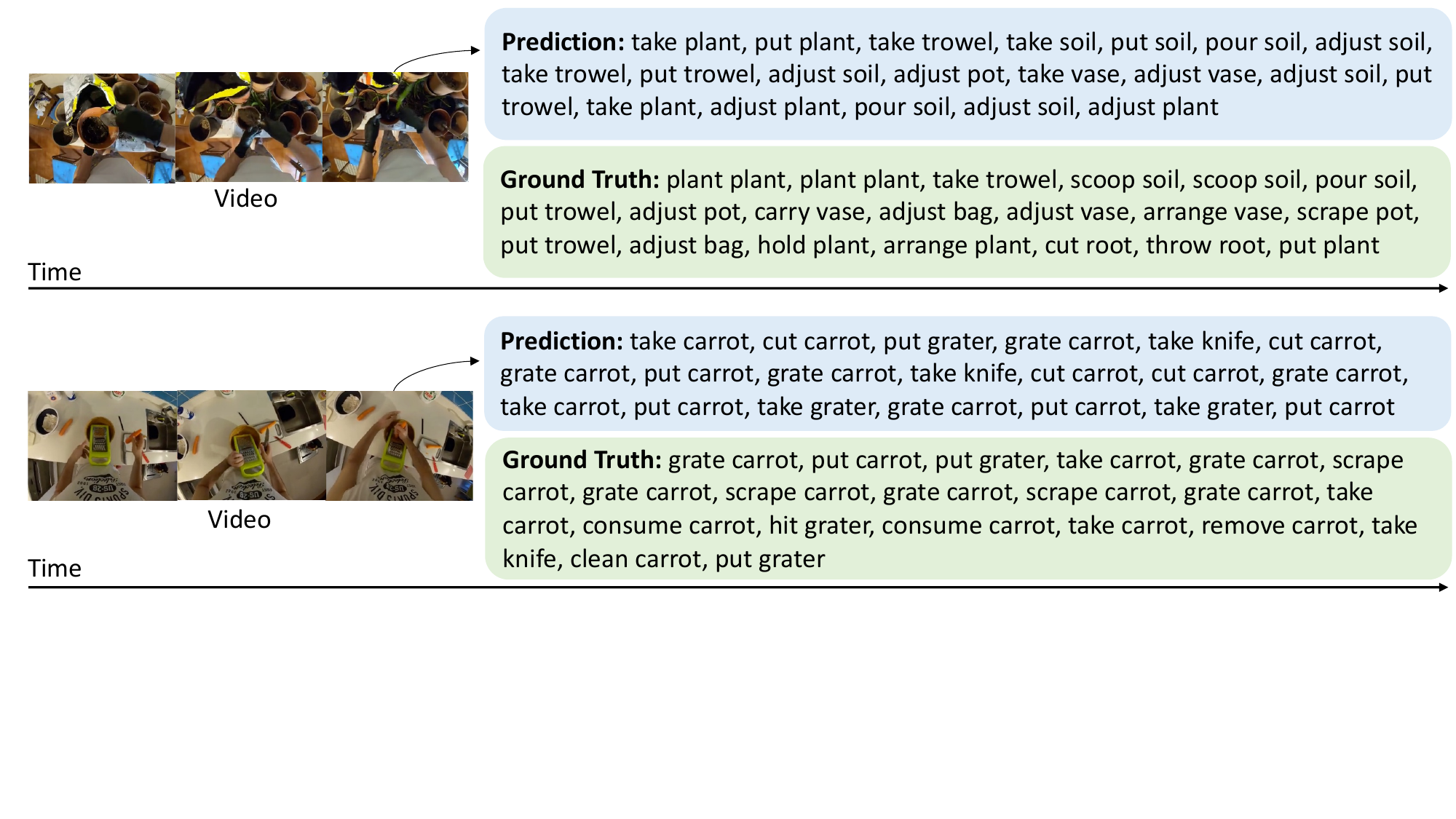}
    \includegraphics[trim=0cm 3.5cm 0cm 0cm, clip, width=0.93\linewidth]{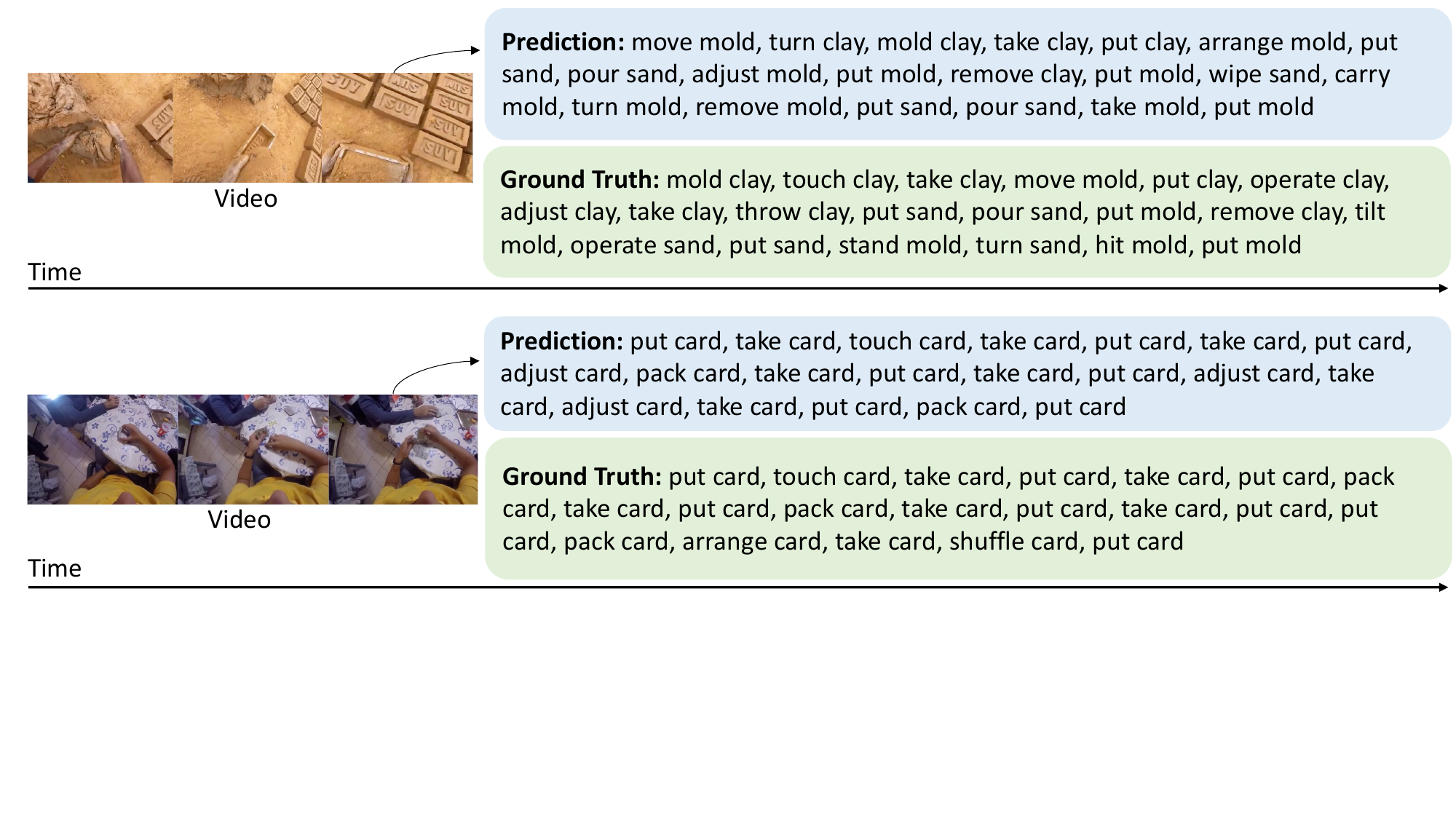}
    \vspace{-6mm}
    \caption{Qualitative Results over videos of diverse environments like kitchen, construction sites, etc. and their respective anticipated actions from our method. Given a video, the top \textcolor{blue}{blue} box shows the prediction from \method and the \textcolor{Green}{green} box contains the ground truth action sequence for reference. The model is able to generate plausible action sequences.}
    \label{fig:supp_qual_results1}
\end{figure*}

\section{Qualitative Analysis}
In this section, we present more qualitative results of our method. Given a video, the top \textcolor{blue}{blue} box shows the prediction from \method and the \textcolor{Green}{green} box contains the ground truth action sequence for reference. We can observe that 
our method is able to understand the activity happening in the video and then, generate action sequences accordingly. Additionally, \method is able to generate action sequences that satisfy the temporal logic constraints and are diverse with less repetitions. The predicted action sequence is also closer to the ground truth action sequence.

\end{document}